%% file: neurips_2023.tex
\documentclass{article}

\usepackage[preprint]{neurips_2023}

\usepackage[utf8]{inputenc} % allow utf-8 input
\usepackage[T1]{fontenc}    % use 8-bit T1 fonts
\usepackage{hyperref}       % hyperlinks
\usepackage{url}            % simple URL typesetting
\usepackage{booktabs}       % professional-quality tables
\usepackage{amsfonts}       % blackboard math symbols
\usepackage{nicefrac}       % compact symbols for 1/2, etc.
\usepackage{microtype}      % microtypography
\usepackage{xcolor}         % colors
\usepackage{enumerate}
\usepackage{enumitem}

\usepackage{hyperref}
\usepackage{url}
\usepackage{multirow}
\usepackage{booktabs}
\usepackage{graphicx}
\usepackage{siunitx}  % For better alignment of numerical values
\usepackage{tabularx}
\usepackage[flushleft]{threeparttable}
\usepackage{enumitem}
\usepackage{amsmath}
\usepackage{xspace}
\usepackage{cleveref}
\usepackage{caption}
\usepackage{subcaption}
\usepackage{pgfplots}
\usepackage{tikz}
\usepackage{wrapfig}
\usepackage{authblk}
\usepackage{fdsymbol}
\usepackage{listings}
\usepackage{float}

\usepackage{adjustbox}
\usepackage{tikz}
\usepackage{pgfplots}
\pgfplotsset{compat=1.18}
\usepackage[edges]{forest}
\definecolor{framework-blue}{RGB}{47, 85, 151}

\definecolor{content-yellow}{RGB}{255, 230, 153}
\definecolor{framework-yellow}{RGB}{255, 255, 255}
\definecolor{content-orange}{RGB}{251, 229, 215}
\definecolor{framework-orange}{RGB}{248, 203, 175}
\definecolor{content-gray}{RGB}{237, 237, 237}
\definecolor{framework-gray}{RGB}{166, 166, 166}

\definecolor{paired-light-blue}{RGB}{198, 219, 239}
\definecolor{paired-dark-blue}{RGB}{49, 130, 188}
\definecolor{paired-light-orange}{RGB}{251, 208, 162}
\definecolor{paired-dark-orange}{RGB}{230, 85, 12}
\definecolor{paired-light-green}{RGB}{199, 233, 193}
\definecolor{paired-dark-green}{RGB}{49, 163, 83}
\definecolor{paired-light-purple}{RGB}{218, 218, 235}
\definecolor{paired-dark-purple}{RGB}{117, 107, 176}
\definecolor{paired-light-gray}{RGB}{217, 217, 217}
\definecolor{paired-dark-gray}{RGB}{99, 99, 99}
\definecolor{paired-light-pink}{RGB}{222, 158, 214}
\definecolor{paired-dark-pink}{RGB}{123, 65, 115}
\definecolor{paired-light-red}{RGB}{231, 150, 156}
\definecolor{paired-dark-red}{RGB}{131, 60, 56}
\definecolor{paired-light-yellow}{RGB}{231, 204, 149}
\definecolor{paired-dark-yellow}{RGB}{141, 109, 49}
\tikzset{%
    parent/.style = {align=center,text width=2.5cm,rounded corners=3pt, line width=0.3mm, fill=gray!10,draw=gray!80},
    child/.style = {align=center,text width=2.3cm,rounded corners=3pt, fill=blue!10,draw=blue!80,line width=0.3mm},
    grandchild/.style = {align=center,text width=2cm,rounded corners=3pt},
    greatgrandchild/.style = {align=center,text width=1.5cm,rounded corners=3pt},
    greatgrandchild2/.style = {align=center,text width=1.5cm,rounded corners=3pt},    
    referenceblock/.style =  {align=center,text width=1.5cm,rounded corners=2pt},
    domain_box/.style= {align=center,text width=2.2cm,rounded corners=3pt, fill=white,draw=framework-blue,line width=0.3mm},
    datasets/.style= {align=center, text width=4.5cm,rounded corners=3pt, fill=paired-light-blue!45,draw=framework-blue,line width=0.3mm},
    few_datasets/.style= {align=center, text width=2.2cm,rounded corners=3pt, fill=paired-light-blue!45,draw=framework-blue,line width=0.3mm},
    models/.style= {align=center, text width=4.5cm,rounded corners=3pt, fill=paired-light-orange!45,draw=framework-blue,line width=0.3mm},
    few_models/.style= {align=center, text width=2.2cm,rounded corners=3pt, fill=paired-light-orange!45,draw=framework-blue,line width=0.3mm},
}
\usepackage{arydshln}

\title{ChatGPT's One-year Anniversary: Are Open-Source Large Language Models Catching up? }

\author{\textbf{Hailin Chen}$^*$$^1$$^,$$^2$,
\textbf{Fangkai Jiao}$^*$$^1$$^,$$^3$,
\textbf{Xingxuan Li}$^*$$^1$,
\textbf{Chengwei Qin}$^*$$^1$,
\textbf{Mathieu Ravaut}$^*$$^1$$^,$$^3$,
\textbf{Ruochen Zhao}$^*$$^1$,
\textbf{Caiming Xiong}$^2$,
\textbf{Shafiq Joty}$^{1,2\dagger}$ \\
$^1$ Nanyang Technological University, Singapore\\
$^2$ Salesforce Research\\
$^3$ Institute of Infocomm Research (I$^{2}$R), A$^{*}$STAR, Singapore\\
{\texttt{\small{\{hailin001, fangkai002, xingxuan001, chengwei003\}@e.ntu.edu.sg}}}\\
{\texttt{\small{\{mathieuj001, ruochen002\}@e.ntu.edu.sg}}}\\
{\texttt{\small{\{cxiong, sjoty\}@salesforce.com}}}
}

\newcommand{\chat}{GPT-3.5-turbo}

\newcommand{\eg}{{\em e.g.,}}

\definecolor{lowyellow}{RGB}{241, 196, 15}

\begin{document}

\maketitle

\def\thefootnote{*}\footnotetext{Authors contributed equally and are ranked by alphabetical order.}\def\thefootnote{\arabic{footnote}}
\def\thefootnote{\dagger}\footnotetext{Work done full-time at Salesforce}\def\thefootnote{\arabic{footnote}}

\begin{abstract}

Upon its release in late 2022, ChatGPT has brought a seismic shift in the entire landscape of AI, both in research and commerce. 
Through instruction-tuning a large language model (LLM) with supervised fine-tuning and reinforcement learning from human feedback, it showed that a model  could answer human questions and follow instructions on a broad panel of tasks.
Following this success, interests in LLMs have intensified, with new LLMs flourishing at frequent interval across academia and industry, including many start-ups focused on LLMs. 
%Notably, LLMs can be split into two groups depending on their availability: open-source and closed-source LLMs. 
%The latter consists of popular chatbot products from tech companies such as OpenAI's GPT-4 or Google's Bard. 
While closed-source LLMs (e.g., OpenAI's GPT, Anthropic's Claude)  generally outperform their open-source counterparts, the progress on the latter has been rapid with claims of achieving parity or even better on certain tasks. This has crucial implications not only on research but also on business. In this work, on the first anniversary of ChatGPT, we provide an exhaustive overview of this success, surveying all tasks where an open-source LLM has claimed to be on par or better than ChatGPT. 

\end{abstract}
\section{Introduction} %hailin
\input{Sections/1_introduction}

\section{Background} \label{sec:Background} % mathieu

In this section, we briefly describe the fundamental concepts that relate to LLMs.

\input{Sections/2_background} 

% ------------------------------------
\section{Open-Source LLMs vs. ChatGPT} \label{sec:main_comparison}

\subsection{General Capabilities} \label{sec:3_1}
% Xingxuan
\input{Sections/3_1_general_capabilities}
% ------------------------------------
% Chengwei
\subsection{Agent Capabilities} \label{sec:3_2}
\input{Sections/3_2_agent}

% \subsubsection{Using Tools}
% \{ToolBench, APIBench, API-Bank, ToolAlpaca, T-Bench\}
% \{gorilla, GSM8K-W-calculator\}
% \subsubsection{Following Feedback}
% \{Mint\}

% \subsubsection{Exploring Environments}
% \{InterCode-CTF, WebArena, M-ALFWorld\} \\
% ------------------------------------
% Fangkai Jiao
\subsection{Problem Solving and Logical Reasoning Capabilities} \label{sec:3_3}

\input{Sections/3_3_logical_reasoning}
% ------------------------------------
% Mathieu
\subsection{Modelling Long-context Capabilities} \label{sec:3_4}
\input{Sections/3_4_long_context} \\
% ------------------------------------
% Hailin
\input{Sections/3_5_application_specific_capabilities} \label{sec:3_5}
% ------------------------------------
% Ruochen
\subsection{Towards Trust-worthy AI} \label{sec:3_6}
\input{Sections/3_6_trustworthy}

% ------------------------------------
\section{Discussion} \label{sec:discussion}
\subsection{Development Trend of LLMs} % Chengwei
\input{Sections/4_1_trend} \label{sec:4_1}

\subsection{Summary of Results} % Ruochen
\input{Sections/4_2_summary} \label{sec:4_2}

\subsection{Recipe of Best Open-source LLMs} % xingxuan
\input{Sections/4_3_recipe} \label{sec:4_3}

\subsection{Loopholes and potential problems} % fangkai
\input{Sections/4_4_problems} \label{sec:4_4}

% ------------------------------------
\section{Conclusion} \label{sec:conclusion}
In this survey, we deliver a systematical review on high performing open-source LLMs that surpass or catch up with ChatGPT in various task domains, at the one-year anniversary mark after ChatGPT's release (\Cref{sec:main_comparison}). In addition, we provide insights, analysis and potential issues of open-source LLMs (\Cref{sec:discussion}). We believe that this survey sheds lights on promising directions of open-source LLMs and will serve to inspire further research and development in the field of open-source LLMs, helping to close the gap with their paying counterparts.

\bibliography{neurips2023_conference}
\bibliographystyle{iclr2024_conference}

\appendix

\end{document}

%% file: Sections/1_introduction.tex
Exactly one year ago, the release of ChatGPT by OpenAI took the AI community and the broader world by storm. For the first time, an application-based AI chatbot could generally provide helpful, safe and detailed answers to most questions, follow instructions, and even admit and fix its previous mistakes. Notably, it can perform these natural language tasks which were traditionally done by  pre-trained then tailored fine-tuned language models such as summarization or question-answering (QA), seemingly amazingly well. As a first of its kind, ChatGPT has attracted the general public -- it reached 100 million users within just two months of its launch, way faster than other popular apps like TikTok or YouTube.\footnote{\href{https://www.reuters.com/technology/chatgpt-sets-record-fastest-growing-user-base-analyst-note-2023-02-01/}{https://www.reuters.com/technology/}} It has also attracted huge  business investments, for its potential to cut down labor cost, automate workflows and even bring new experiences to customers~\citep{gpt-data-analyst}.
% \sj{cite}. 

However, since ChatGPT is not open-sourced and its access is controlled by a private company, most of its technical details remain unknown. Despite the claim that it follows the procedure introduced in InstructGPT (also called GPT-3.5) \citep{instructgpt}, its exact architecture, pre-training data and fine-tuning data are unknown. Such close-source nature generates several key issues. First, without knowing the internal details such as the pre-training and fine-tuning procedure, it is hard to properly estimate its potential risks to the society, especially knowing that LLMs can notoriously  generate toxic, unethical and untruthful content. Second, it has been reported that ChatGPT's performance changes over time hindering reproducible results \citep{chen2023chatgpt}. Third, ChatGPT has experienced multiple outages, with two major ones only in November 2023 during which the access to ChatGPT website and its API was completely blocked. Finally, enterprises adopting ChatGPT may be concerned with the heavy cost of calling APIs, service outages, data ownership and privacy issues, and other unpredictable events such as the recent boardroom drama about the CEO Sam Altman's dismissal to staff rebellion, and his eventual return (REUTERS \href{https://www.reuters.com/technology/openai-ouster-microsoft-ai-research-ceo-sam-altmans-tumultuous-weekend-2023-11-20/}{source}).

%to research community, its changing underlying model could cause research papers evaluating or using ChatGPT become un-reproducible or even conflicting with each other.
%  may not stay constant over time, with major loss of performance on some evaluation benchmarks after a model update \citep{chen2023chatgpt}. 
  
\input{Sections/latex_figures/general_radar_charts}

Open-source LLMs, on the other hand, offer a promising direction as they can potentially remediate or bypass most of the aforementioned issues. For this reason, the research community has been actively pushing for maintaining high-performing LLMs in open-source.  However, as it stands today (as of late 2023), it is widely believed that open-source LLMs such as Llama-2 \citep{llama2} or Falcon \citep{falcon40b} lag behind their closed-source counterparts such as OpenAI's GPT3.5 (ChatGPT) and GPT-4 \citep{gpt4}, Anthropic's Claude\footnote{\href{https://www.anthropic.com/index/introducing-claude}{https://www.anthropic.com/index/introducing-claude}} or Google's Bard\footnote{\href{https://blog.google/technology/ai/bard-google-ai-search-updates/}{https://blog.google/technology/ai/bard-google-ai-search-updates}}, with GPT-4 generally assumed to champion them all. However, what is very encouraging is that the gap is getting narrower and narrower, and open-source LLMs are quickly catching up. In fact, as it is illustrated in \Cref{fig:radar}, the best open-source LLMs already perform better than \chat\ on some standard benchmarks. Yet, it is not a straightforward uphill battle for open-source LLMs. The landscape is constantly evolving: closed-source LLMs are updated by retraining on newer data regularly, open-source LLMs are released to catch up, and there is a myriad of evaluation datasets and benchmarks being used to compare LLMs, making singling out a best LLM especially challenging.

% final paragraph
In this survey, we aim to consolidate recent studies on open-source LLMs and provide an overview of open-source LLMs that match or surpass ChatGPT in various domains. Our contributions are three-fold:
\begin{itemize}[leftmargin=*,topsep=3pt,itemsep=3pt,parsep=0pt]
    \item Consolidating various evaluations of open-source LLMs, providing an unbiased and comprehensive view of open-source LLMs vs. ChatGPT (\Cref{fig:radar}, \Cref{sec:3_1}).
    
    \item Systematically reviewing open-source LLMs that match or surpass the performance of  ChatGPT in various tasks with analysis (\Cref{fig:domain_tree}, \Cref{sec:main_comparison}, \Cref{sec:4_2}). We are also maintaining a live web page to track the latest updates.\footnote{ \href{https://github.com/ntunlp/OpenSource-LLMs-better-than-OpenAI/tree/main}{https://github.com/ntunlp/OpenSource-LLMs-better-than-OpenAI/tree/main}}
    
    \item Presenting insights on the trend of open-source LLMs development (\Cref{sec:4_1}), the good practices to train open-source LLMs (\Cref{sec:4_3}) and potential issues with open-source LLMs (\Cref{sec:4_4}).
\end{itemize}

\paragraph{Who can benefit from this survey?}  This survey aims to serve as a pivotal resource for both the research community and business sector in understanding the current landscape and future potential of open-source LLMs. For researchers, it provides a detailed synthesis of the current progress and evolving trends in open-source LLMs, highlighting promising directions for future investigation. For the business sector, this survey offers valuable insights and guidance, assisting decision-makers in evaluating the applicability and benefits of adopting open-source LLMs.

In the following, we start by introducing background concepts (\Cref{sec:Background}), then provide an in-depth review of open-source LLMs that beat ChatGPT in various domains (\Cref{sec:main_comparison}), followed by a discussion on insights and issues of open-source LLMs (\Cref{sec:discussion}), finally we conclude with a summary  (\Cref{sec:conclusion}). 

\input{Sections/latex_figures/domain_tree}

%% file: Sections/latex_figures/general_radar_charts.tex
\begin{figure}[t]
    \centering
    \begin{subfigure}[t]{.48\textwidth}
        \centering
        \includegraphics[width=\linewidth]{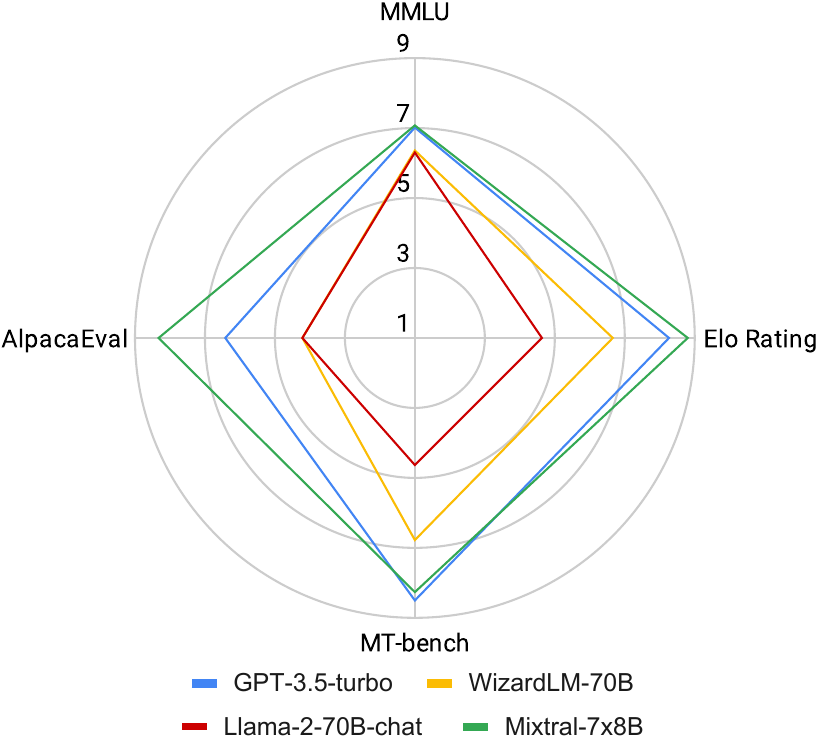}
        \caption{Open-source LLMs vs. ChatGPT on MT-bench \cite{mtbench}, ELO Rating, MMLU \citep{mmlu} and AlpacaEval \cite{alpacaeval}.} %WizardLM-70B is not evaluated on AlpacaEval, we take the performance of WizardLM-13B-v1.2 as its estimated performance.
        \label{subfig:radar1}
    \end{subfigure}
    \hfill
    \begin{subfigure}[t]{.48\textwidth}
        \centering
        \includegraphics[width=1.0\linewidth]{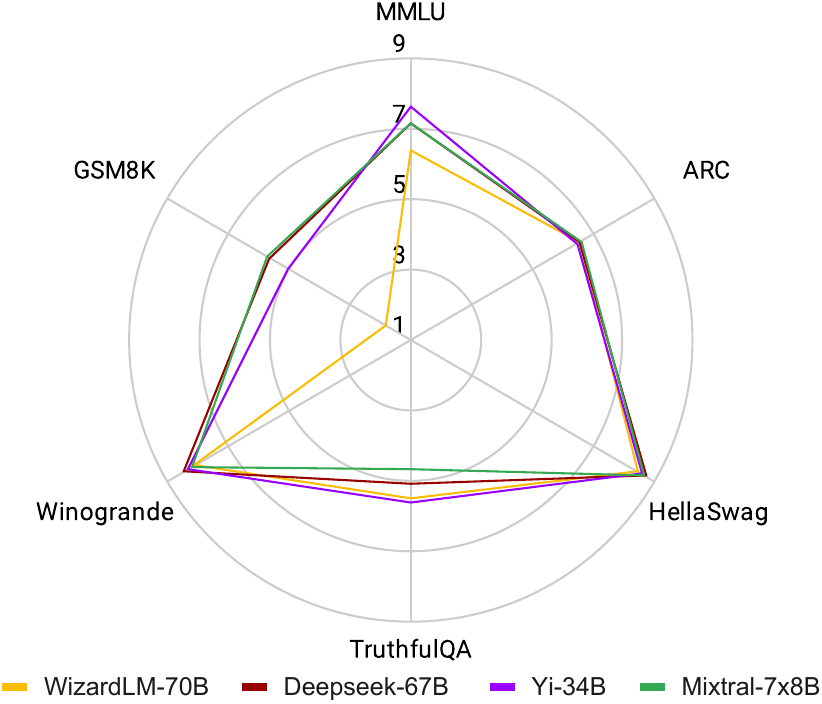}
        \caption{Selected open-source LLMs on HuggingFace's Open LLM leaderboard.}
        \label{subfig:radar2}
    \end{subfigure}
    \caption{Overview of different open-source LLMs on various general benchmarks.}
    \label{fig:radar}
\end{figure}

%% file: Sections/latex_figures/domain_tree.tex
\begin{figure*}[t]
\scriptsize
    \begin{adjustbox}{width=\textwidth}
        \begin{forest}
        for tree={
                forked edges,
                grow'=0,
                draw,
                rounded corners,
                node options={align=center},
                text width=2.7cm,
                s sep=6pt,
                calign=edge midpoint, 
            },
                [LLM's Capabilities, fill=gray!45, parent
                    [General \S\ref{sec:3_1}, domain_box
                        [{AlpacaEval, MT-bench, ELO rating, Open-LLM leaderboard}, datasets
                            [{Llama-2, WizardLM, Zephyr, Deepseek, Yi, Mixtral}, models]
                        ]
                    ]
                    [Agent \S\ref{sec:3_2}, domain_box
                        [Using Tools, domain_box
                            [{API-Bank, ToolBench, APIBench, ToolAlpaca, etc.}, datasets
                                [{Gorilla, ToolLLaMA}, few_models]
                            ]
                        ]
                        [Self-Debugging, domain_box
                            [{InterCode-Bash, InterCode-SQL, MINT-MBPP, MINT-HumanEval, RoboCodeGen, etc.}, datasets
                            ]
                        ]
                        [Following NL Feedback, domain_box
                            [{MINT}, few_datasets]
                        ]
                        [Exploring Environments, domain_box
                            [{ALFWorld, InterCode-CTF, WebArena}, datasets]
                        ]
                        [{Lemur-chat, AgentLlama, OpenChat-3.5}, models]
                    ]
                    [Logical Reasoning \S\ref{sec:3_3}, domain_box
                        [Math, domain_box
                            [{GSM8K, MATH, TheoremQA, etc.}, datasets
                                [{WizardMath}, few_models]
                            ]
                        ]
                        [Coding, domain_box
                            [{HumanEval, MBPP, APPs, etc.}, datasets
                                [{WizardCoder}, few_models]
                            ]
                        ]
                    ]
                    [Modelling Long-context \S\ref{sec:3_4}, domain_box
                        [{SCROLLS, Zero-SCROLLS, LongBench, L-Eval, BAMBOO, M4LE, etc.}, datasets
                            [{Llama-2-long}, few_models]
                        ]
                    ]
                    [Application-specific \S\ref{sec:3_5}, domain_box
                        [Query-focused Summarization, domain_box
                            [{QMSum, SQuALITY, CovidET, NEWTS}, datasets
                                [{\textit{finetuned} model}, few_models]
                            ]
                        ]
                        [Open-ended QA, domain_box
                            [{NQ,TriviaQA,NewsQA,SQuAD,Quoref,
                            NarrativeQA,DROP}, datasets
                                [{InstructRetro}, few_models]
                            ]
                        ]
                        [Medical, domain_box
                            [{Dreaddit, loneliness, MIMIC-CXR, OpenI, etc.}, datasets
                                [{MentaLLaMA, Radiology-Llama-2}, few_models]
                            ]
                        ]
                        [Generating Structured Data, domain_box
                            [{Rotowire, Struc-Bench-Latex, Struc-Bench-HTML}, datasets
                                [{Struct-Bench}, few_models]
                            ]
                        ]
                        [Generating Critiques, domain_box
                            [{AlpacaFarm, FairEval, CritiqueEval, etc.}, datasets
                                [{Shepherd}, few_models]
                            ]
                        ]
                    ]
                    [Trustworthiness \S\ref{sec:3_6}, domain_box
                        [Hallucination, domain_box
                            [{TruthfulQA, FactualityPrompt, FActScore, 
                            KoLA-KC, HaluEval, FACTOR}, datasets
                                [{Platypus, 
                                Chain-of-Verification, etc.}, few_models]
                            ]
                        ]
                        [Safety, domain_box
                            [{SafetyBench, XSTEST, etc.}, datasets]
                        ]
                    ]
                ] 
        \end{forest}
    \end{adjustbox} 
    \caption{Typology of LLM's capabilities and best performing open-LLMs. White boxes denote domains, \colorbox{paired-light-blue}{blue boxes} represent specific datasets and \colorbox{paired-light-orange}{orange boxes} denote open-sourced LLMs.}
    \label{fig:domain_tree}
\end{figure*}
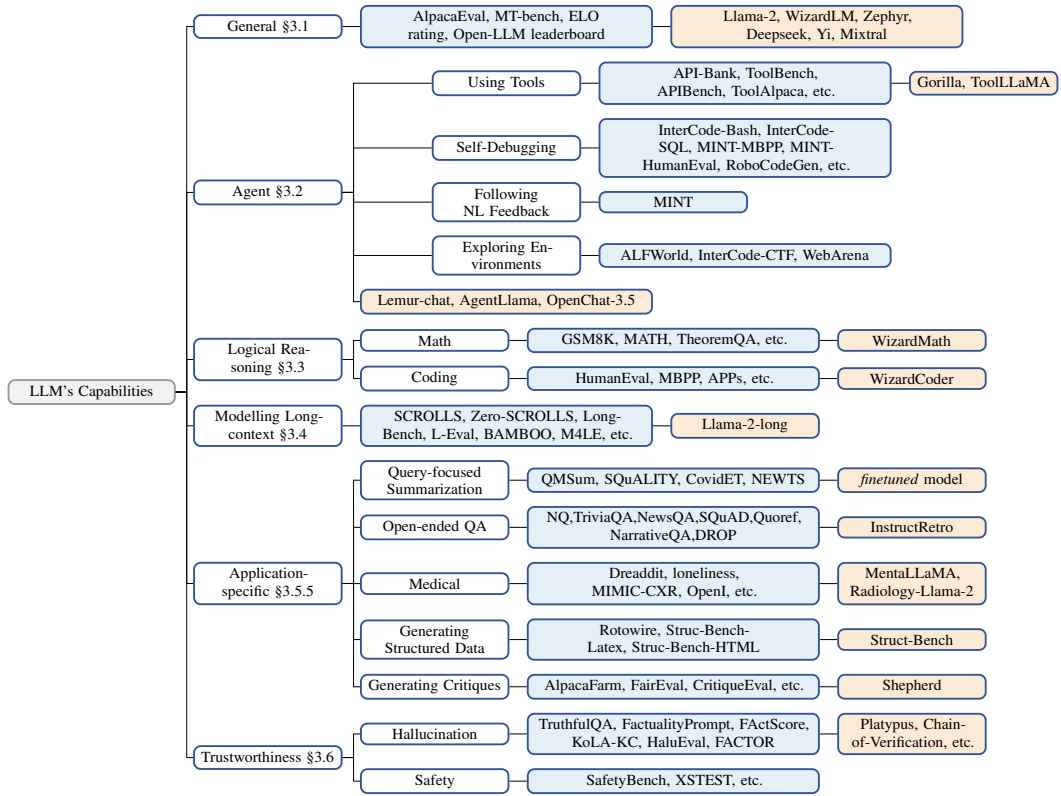

%% file: Sections/2_background.tex
\subsection{Training Regimes}

\paragraph{Pre-training}
All LLMs rely on large-scale self-supervised pre-training on Internet text data \citep{gpt,gpt3}. Decoder-only LLMs follow the causal language modeling objective, through which the model learns to predict the next token conditioning on the sequence of previous tokens \citep{bengio2000neural}. As per pre-training details shared by open-source LLMs \citep{llama}, sources of text data include CommonCrawl\footnote{\href{https://commoncrawl.org/}{https://commoncrawl.org}}, C4 \citep{t5}, GitHub, Wikipedia, books, and online discussion exchanges such as Reddit or StackOverFlow. It is widely acknowledged that scaling the size of pre-training corpus improves the model performance, and works hand-in-hand with scaling the model size, a phenomenon referred to as \emph{scaling laws}, and analyzed in depth in \citep{chinchilla}. Modern-day LLMs pre-train on a corpus from hundreds of billions to several trillions of tokens \citep{llama2, falcon}. 

\paragraph{Fine-tuning}
Fine-tuning aims to adapt a pre-trained LLM to downstream tasks, by updating weights with the available supervision, which usually forms a dataset orders of magnitude smaller than the one used for pre-training \citep{bert}. T5 \citep{t5} was among the first to frame fine-tuning into a text-to-text unified framework, with natural language instructions describing each task. \textbf{Instruction-tuning} later extended fine-tuning by training jointly on several tasks \citep{instructiontuning, ext5}, each described with natural language instructions. Instruction-tuning quickly gained in popularity, due to its ability to drastically improve zero-shot performance of LLMs, including on new tasks (unseen during training), and especially at larger models scale. Standard instruction-tuning with multi-task supervised fine-tuning (commonly known as \textbf{SFT}) may still not result in models that follow humans intentions while being safe, ethical and harmless, and can be futher improved with Reinforcement Learning from Human Feedback (\textbf{RLHF}): human annotators rank outputs from the fine-tuned model, which are used to fine-tune again with reinforcement learning \citep{instructgpt}. Recent work showed that human feedback may be replaced with feedback from an LLM, a process referred to as Reinforcement Learning from AI Feedback (\textbf{RLAIF}) \cite{bai2022constitutional}. 
Direct Preference Optimization (\textbf{DPO}) bypasses the need to fit a reward model to human preferences as in RLHF and instead directly fine-tunes the policy with a cross-entropy objective, achieving more efficient alignment of the LLM to human preferences. 
%\sj{Introduce DPO here}

A line of work focuses on \emph{quality over quantity} when building an instruction-tuning dataset of diverse tasks: Lima \citep{lima} outperforms GPT-3 with a Llama-65B fine-tuned on just 1,000 examples, and Alpagasus \citep{alpagasus} improves on Alpaca \citep{alpaca} by cleaning its instruction fine-tuning dataset from 52k to 9k examples.

\paragraph{Continual pre-training}
Continual pre-training consists in performing another round of pre-training from a pre-trained LLM, typically with a lesser volume of data than in the first stage. Such process may be useful to quickly adapt to a new domain or elicit new properties in the LLM. For instance, continual pre-training is used in Lemur \citep{lemur} to improve coding and reasoning capacities, and Llama-2-long \citep{llama2long} to extend context window.
%For instance, Lemur \citep{lemur} adapts Llama-2 to the code domain by continual pre-training on a 90B tokens dataset. Llama-2-long \citep{llama2long} adapts the base Llama-2 \citep{llama2} to longer sequences by continuing pre-training on 400B tokens with context length 16k tokens (up from 4k in the base model).

\paragraph{Inference}
There exists several alternative methods for sequence generation with auto-regressive decoding with an LLM, which differ by the degree of randomness and diversity in the output. Increasing the temperature during sampling makes outputs more diverse, while setting it to 0 falls back to greedy decoding, which may be needed in scenarios necessitating deterministic outputs. Sampling methods top-k \citep{topk} and top-p \citep{topp} constrain the pool of tokens to sample from at each decoding step.

Several techniques aim to improve inference speed, especially at longer sequence length, which become problematic due to the attention complexity, which is quadratic with regards to input length. FlashAttention \citep{flashattention} optimizes reads/writes between levels of GPU memory, accelerating both training and inference. FlashDecoding \citep{flashdecoding} parallelizes the key-value (KV) cache loading in the attention mechanism, yielding a 8x end-to-end speedup. Speculative decoding \citep{speculativedecodinggoogle,speculativedecodingdeepmind} uses an extra, small language model to approximate next token distribution from an LLM, which accelerates decoding without loss of performance. vLLM \citep{vllm} accelerates LLM inference and serving using PagedAttention, an algorithm for optimizing memory usage of attention keys and values. 
% \sj{Could we briefly mention efficient inference methods (flash decoding, speculative, page attention, flash attention, etc.) and library like vLLM}

\subsection{Task Domains and Evaluation}

Properly assessing the capabilities of LLMs remains an active research area, due to the diversity and breadth of evaluations to perform. Question-answering datasets \citep{TrivialQA, NaturalQuestions, truthfulqa} are very popular evaluation benchmarks, but new benchmarks tailored for LLM assessments have also emerged recently \citep{alpacafarm, open-llm-leaderboard, mtbench}. In the following section, we explore LLMs capacities across 6 main dimensions: general capabilities, agent capabilities, logical reasoning (including maths and coding capacities), long-context modelling, specific applications such as QA or summarization, and trustworthiness. 

%% file: Sections/3_1_general_capabilities.tex
% \{MT-bench, AlpacaEval, open\_llm\_leaderboard\} and \{BigBench, ChatEval, MMLU?\}
\paragraph{Benchmarks}   
As numerous LLMs are released week upon week, each claiming superior performance on certain tasks, it becomes increasingly challenging to identify true advancements and the leading models. Therefore, it is crucial to comprehensively assess the performance of these models across a broad spectrum of tasks to understand their general capabilities. This section covers benchmarks using LLM-based (\textit{e.g.,} GPT-4) evaluation and traditional (\textit{e.g.,} ROUGE \citep{rouge} and BLEU \citep{bleu}) evaluation metrics.

% \sj{itemize them}
\begin{itemize}[leftmargin=*,topsep=3pt,itemsep=3pt,parsep=0pt]
    \item \textbf{MT-Bench} \citep{mtbench} is designed to test multi-turn conversation and instruction-following ability from eight perspectives:  writing, roleplay, information extraction, reasoning, math, coding, knowledge I (STEM), and knoweldge II (humanities/social science). Stronger LLMs (\textit{e.g.,} GPT-4) are utilized as judges to evaluate the models for this benchmark.
    \item \textbf{AlpacaEval} \citep{alpacaeval} is an LLM-based automatic evaluator based on AlpacaFarm \citep{alpacafarm} evaluation set, which tests the ability of models to follow general user instructions. It benchmarks candidate models against Davinci-003 responses utilizing stronger LLMs (\textit{e.g.,} GPT-4 and Claude), which generate the candidate model's win rate.
    \item \textbf{Open LLM Leaderboad} \citep{open-llm-leaderboard} evaluates LLMs on seven key benchmarks using the Language Model Evaluation Harness \citep{eval-harness}, including AI2 Reasoning Challenge \citep{arc}, HellaSwag \citep{hellaswag}, MMLU \citep{mmlu}, TruthfulQA \citep{truthfulqa}, Winogrande \citep{winogrande}, GSM8K \citep{gsm8k}, and DROP \citep{drop}. This framework evaluates LLMs on a variety of reasoning and general knowledge across a wide variety of fields in zero-shot and few-shot settings.
    \item \textbf{BIG-bench} \citep{bigbench} is a collaborative benchmark aimed to probe LLMs and extrapolate their future capabilities.
    It includes more than 200 novel language tasks, covering a diverse range of topics and languages, which are not entirely solvable by existing models.
    \item \textbf{ChatEval} \citep{chateval} is a multi-agent debate framework, which enables a multi-agent referee team to autonomously discuss and evaluate the quality of generated responses from different models on open-ended questions and traditional natural language generation tasks. 
    \item \textbf{FairEval-Vicuna} \citep{FairEval} utilizes both multiple evidence calibration and balanced position calibration on a set of 80 questions from the Vicuna Benchmark \citep{mtbench}. FairEval-Vicuna offers a more impartial evaluation outcome within the paradigm of adopting LLMs as evaluators, which closely aligns with human judgements.
\end{itemize}

\begin{wraptable}{r}{0.6\textwidth}
    \centering
    \vspace{-1.0em}
    \resizebox{0.6\textwidth}{!}{
        \begin{tabular}{lccc}
            \toprule
            
            \textbf{Models}           
            & \textbf{MT-Bench} 
            & \textbf{AlpacaEval-2.0} 
            & \textbf{Open LLM Leaderboard} \\
            
            \midrule
            
            \textbf{Llama-2-70B-chat} & 6.86 & 13.87 & - \\
            \textbf{WizardLM-70B}     & 7.71 & 12.03 & 57.17 \\
            \textbf{Zephyr-7B}        & 7.34 & 10.99 & 52.15 \\
            \textbf{Yi-34B}           & - & \textbf{29.66} & 69.42 \\
            \textbf{Mixtral-8x7B}     & 8.30 & 18.26 & 68.42 \\
            
            \midrule
            
            \textbf{GPT-3.5-turbo}    & \underline{8.39} & 14.13 & \underline{70.21} \\
            \textbf{GPT-4}            & \textbf{8.99} & \underline{23.58} & \textbf{85.36} \\
            
            \bottomrule                        
        \end{tabular}
    }
    \caption{Model performance on general benchmarks.}
    \vspace{-0.5em}
\end{wraptable}

\paragraph{Performance of LLMs}
Llama-2-70B \citep{llama2}, a prominent open-source LLM from Meta, has been pre-trained on a massive dataset of two trillion tokens. It demonstrates remarkable results across various general benchmarks.
%When further fine-tuned with instruction data, the Llama-2-chat-70B variant exhibits enhanced capabilities in general conversational tasks.
In particular, Llama-2-chat-70B achieves a 13.87\% win rate in AlpacaEval-2.0, approaching the performance of \chat~. %Nonetheless, GPT-4 remains the top performer among all LLMs with a win rate of 95.28\%. 
Zephyr-7B \citep{zephyr}, a much smaller model, uses distilled direct preference optimization \citep{dpo} and achieves better results to 70B Llama-2 on MT-bench, scoring 7.34 against 6.86. 
%It even surpasses Llama-2-chat-70B on MT-Bench, scoring 7.34 against 6.86. 
Additionally, WizardLM-70B \citep{wizard-lm} has been instruction fine-tuned using large amounts of instruction data with varying levels of complexity. It achieves a good score of 7.71 on MT-Bench. %However, this is still slightly lower than the scores of \chat~(7.94) and GPT-4 (8.99).
Although Zephyr-7B shows top performance in the MT-Bench, it falls short in the Open LLM Leaderboard, scoring only 52.15\%. 

%On the other hand, GodziLLa2-70B \citep{godzila}, an experimental model that combines various proprietary LoRAs from Maya Philippines \footnote{Maya (\href{https://www.maya.ph}{https://www.maya.ph}) is a Filipino financial services and digital payments company.} and the Guanaco Llama 2 1K dataset \citep{guanacollama2} with Llama-2-70B, achieves a more competitive score of 67.01\% on the Open LLM Leaderboard. 
Furthermore, Yi-34B pre-trained from scratch by developers at 01.AI \footnote{\href{https://www.01.ai}{https://www.01.ai}}, stands out among all open-source LLMs with a remarkable score of 69.42\% on .
This performance is comparable to that of \chat, which scores 70.21\%. Moreover, Yi-34B achieves a stunning score of 29.66\% on AlpacaEval-2.0, surpassing GPT-4 with a significant margin. Mixtral\citep{mixtral}, a mixture of experts language model, is another top-performing open-source pretrained LLM, approaching \chat~ performance on MT-Bench and Open LLM Leaderboard while outperforming \chat~ on AlpacaEval-2.0. %However, both are still notably behind GPT-4, which leads with a substantial score of 85.36\%. 

Besides, UltraLlama \citep{ultrallama} utilizes fine-tuning data with enhanced diversity and quality. It matches \chat's performance in its proposed benchmark while exceeding it in areas of world and professional knowledge. Deepseek-67B\citep{deepseek}, a recently released pretrained LLM with 15 more layers than Llama-2-70B, has demonstrated to surpass GPT-3.5-turbo on AlignBench\citep{alignbench} -- a comprehensive LLM benchmark in Chinese.

% \sj{Do you want to refer back to Fig 1 and also include Yi in Table 1?}

% \paragraph{Inference-time techniques}

% \subsubsection{Pretrained/finetuned Open-LLMs}
% \subsubsection{Inference-time techniques with Open-LLMs}

%% file: Sections/3_2_agent.tex
\paragraph{Benchmarks} With the recent advancements in scaling up model size, LLM-based agents (also called \emph{language agents}) have drawn a great deal of attention from the AI community. In light of this, we investigate the agent capabilities of open-source LLMs on a variety of benchmarks. Depending on the skills required, existing benchmarks can be mainly divided into four categories.

\begin{itemize}[leftmargin=*,topsep=3pt,itemsep=3pt,parsep=0pt]

\item \emph{Using Tools}: Some benchmarks have been proposed to evaluate the tool usage capabilities of LLMs. \textbf{API-Bank} \citep{li2023api} is specifically designed for tool-augmented LLMs. \textbf{ToolBench} \citep{xu2023tool} is a tool manipulation benchmark including various software tools for real-world tasks. \textbf{APIBench} \citep{gorilla} consists of APIs from HuggingFace, TorchHub, and TensorHub. \textbf{ToolAlpaca} \citep{toolalpaca} develops a diverse and comprehensive tool-use dataset through a multi-agent simulation environment. Coincidentally, another instruction-tuning dataset constructed using ChatGPT for tool use is also named \textbf{ToolBench} \citep{toolllm}.  Besides, \textbf{MINT} \citep{mint} can evaluate the proficiency of LLMs in employing tools to solve tasks that necessitate multi-turn interactions.
%\textbf{AgentBench} includes 8 different environments for assessing the reasoning and decision-making abilities of LLMs.

\item \emph{Self-Debugging}: Several datasets are available to assess the ability of LLMs to self-debug, including \textbf{InterCode-Bash} and \textbf{InterCode-SQL} \citep{intercode}, \textbf{MINT-MBPP} and \textbf{MINT-HumanEval} \citep{mint}, and \textbf{RoboCodeGen} \citep{liang2023code}.

\item \emph{Following Natural Language Feedback}: \textbf{MINT} \citep{mint} can also be used to measure the ability of LLMs to leverage natural language feedback by using GPT-4 \citep{gpt4} to simulate human users.

\item \emph{Exploring Environment}: \textbf{ALFWorld} \citep{alfworld}, \textbf{InterCode-CTF} \citep{intercode}, and \textbf{WebArena} \citep{webarena} are introduced to evaluate whether LLMs-based agents are able to gather information from the environment and make decisions. 

\end{itemize}

\paragraph{Performance of LLMs} By pre-training Llama-2 using a code-intensive corpus containing 90B tokens and instruction fine-tuning on 300K examples including both text and code, Lemur-70B-chat \citep{lemur} surpasses the performance of \chat~when exploring the environment or following natural language feedback on coding tasks. 
\begin{wraptable}{r}{0.6\textwidth}
    \centering
    \vspace{-1.0em}
    \resizebox{0.6\textwidth}{!}{
        \begin{tabular}{@{}lccccc@{}}
            \toprule
            
            \multirow{2}{*}{\textbf{Model}} 
            & \multicolumn{3}{c}{\textbf{Environment}} &  \textbf{NL Feedback} \\ 
            \cmidrule(l){2-5} 
            & \textbf{ALFWorld}   
            & \textbf{IC-CTF}  
            & \textbf{WebArena} 
            & \textbf{Code Generation} \\ 
            
            \midrule
            
            \textbf{Lemur-70B-chat}   & \underline{59.70} & \underline{22.00} & 5.30 & \textbf{17.65} \\ 
            
            \midrule
            
            \textbf{\chat}            & 41.79 & 11.00 & \underline{7.38} & \underline{9.56} \\
            \textbf{GPT-4}            & \textbf{84.33} & \textbf{37.00} & \textbf{10.59} & - \\ 
             
            \bottomrule
        \end{tabular}
    }
    \caption{Model performance on several agent benchmarks.}
    \vspace{-0.5em}
\end{wraptable}
AgentTuning \citep{agenttuning} conducts instruction tuning with Llama-2 on the combination of its constructed AgentInstruct dataset and general domain instructions, resulting in AgentLlama. Notably, AgentLlama-70B achieves comparable performance to \chat~on unseen agent tasks. Through fine-tuning Llama-2-7B on ToolBench, ToolLLaMA \citep{toolllm} demonstrates comparable performance to \chat~in tool usage evaluations. \citet{chen2023fireact} introduce FireAct, which can fine-tune Llama-2-13B to outperform prompting \chat~on HotpotQA \citep{yang2018hotpotqa}. Gorilla \citep{gorilla}, fine-tuned from Llama-7B, outperforms GPT-4 on writing API calls. More recently, OpenChat-3.5\citep{openchat}, an LLM pretrained with mixed quality data,  has demonstrated to outperfom GPT-3.5-turbo on several agentic tasks, according to \cite{pangu-agent}.

%% file: Sections/3_3_logical_reasoning.tex
\paragraph{Benchmarks}

Problem solving and logical reasoning serve as fundamental capability of high-level ability and skill, such as programming, theorem proving, as well as arithmetic reasoning. To this end, in this section, we will cover the following benchmarks:

% \sj{itemize them}
\begin{itemize}[leftmargin=*,topsep=3pt,itemsep=3pt,parsep=0pt]
    \item \textbf{GSM8K}~\citep{gsm8k} consists of 8.5K high quality grade school math problems created by human problem writers. These problems take between 2 and 8 steps to solve, and solutions primarily involve performing a sequence of elementary calculations using basic arithmetic operations to reach the final answer.
    \item \textbf{MATH}~\citep{math2021} is a dataset of 12,500 challenging competition mathematics problems. Each problem in MATH has a full step-by-step solution which can be used to teach models to generate answer derivations and explanations.
    \item \textbf{TheoremQA}~\citep{theoremqa} is a theorem-driven QA dataset designed to evaluate AI models’ capabilities to apply theorems to solve challenging science problems. It was curated by domain experts containing 800 high-quality questions covering 350 theorems from Math, Physics, EE\&CS, and Finance.
    \item \textbf{HumanEval}~\citep{codex} is a set of 164 hand written programming problems. Each problem includes a function signature, docstring, body, and several unit tests, with an average of 7.7 tests per problem. 
    \item \textbf{MBPP}~\citep{mbpp} (The Mostly Basic Programming Problems) dataset contains 974 short Python programs constructed by crowd-sourcing to an internal pool of crowd workers who have basic knowledge of Python. Each problem is assigned with a self-contained Python function solving the problem specified, and three test cases that check for semantic correctness of the function.
    \item \textbf{APPs}~\citep{apps} is a benchmark for code generation measuring the ability of models to take an arbitrary natural language problem specification and generate satisfactory Python code. The benchmark includes 10,000 problems, which range from having simple one line solutions to being substantial algorithmic challenges.
\end{itemize}

\begin{wraptable}{r}{0.45\textwidth}
    \centering
    \vspace{-0.4cm}
    \renewcommand{\arraystretch}{1.2}
    \resizebox{0.45\textwidth}{!}{
        \begin{tabular}{lcc}
            \toprule
            
            \textbf{Models}  
            & \textbf{GSM8K} 
            & \textbf{HumanEval}\\ 
        
            \midrule
            
            \textbf{WizardMath-7B}   & 54.9 & - \\
            \textbf{WizardMath-13B}  & 63.9 & - \\
            \textbf{WizardMath-70B}  & \underline{81.6} & - \\ 
            \textbf{WizardCoder-15B} & - & 57.3 \\
            
            \hdashline
            
            \textbf{Lemur-70B}       & 54.9 & 35.4 \\
            \textbf{Lemur-70B-chat}  & 66.3 & 61.0 \\
            
            \hdashline
            
            \textbf{OpenChat-3.5}    & 71.3 & \textbf{77.4} \\
            
            \hdashline
            
            \textbf{Phi-1-1.3B}      & - & 50.6 \\
            \textbf{Phi-1.5-1.3B}    & - & 41.4 \\

            \midrule
            
            \textbf{\chat}           & 57.1 & 48.1 \\
            \textbf{GPT-4}           & \textbf{92.0} & \underline{67.0} \\
            
            \bottomrule
        \end{tabular}
    }
    \caption{Model Performance on two benchmarks requiring logical reasoning, GSM8K (math) and HumanEval (coding).}
    \vspace{-0.5cm}
    \label{tab:logicl-reasoning}
\end{wraptable}

% \sj{We should try to put the numbers and models in a table}

\paragraph{Enhanced Instruction Tuning} Different from conventional knowledge distillation based instruction tuning, \cite{wizard-code,wizard-math} employed \textit{Evol-Instruct}~\citep{wizard-lm} to construct a task-specific high quality instruction tuning dataset, where the seed instructions from Alpaca~\citep{alpaca} have evolved to the ones either extended in knowledge boundary or the depth of task complexity. After obtaining the expanded instruction pool, the new instruction tuning dataset is generated by collecting responses from another LLM, e.g., \chat.
Besides, \cite{wizard-math} also incorporate PPO~\citep{ppo} algorithm to further improve the quality of both generated instruction and answer. Specifically, they first train an instruction reward model (IRM) to rank the quality of generated evolved instructions, as well as a process-supervised reward model (PRM) to provide process supervision to intermediate reasoning process. The quality supervision is generated from a fine-tuned Llama model while the process supervision come from \chat.
Finally, benefiting from the evolved depth and width of queries, the fine-tuned student model achieves even better performance than \chat. For example, WizardCoder~\citep{wizard-code} outperforms \chat~on HumanEval with 19.1\% absolute improvements. WizardMath~\citep{wizard-math} has also obtained 42.9\% absolute improvements on GSM8K compared with \chat.

\paragraph{Pre-training on Data with Higher Quality} Lemur~\citep{lemur} has verified a better mixture of natural language data and code and induces the LLMs with stronger abilities on function calling, automatic programming, and agent capabilities. Specifically, Lemur-70B-chat achieves significant improvements over \chat~on both HumanEval and GSM8K without task-specific fine-tuning. Phi-1 and Phi-1.5~\citep{phi-1,phi-15} take a different road by using the textbook as the main corpus for pre-training, which makes the strong abilities observable on much smaller language models. Furthermore, OpenChat-3.5 outperforms GPT-3.5-turbo on both GSM8K and HumanEval witha significant margin, with only 7B parameters.

%% file: Sections/3_4_long_context.tex
\paragraph{Benchmarks} Processing long sequences remains one of the key technological bottlenecks of LLMs, as all models are limited by a finite maximum context window, typically from 2k to 8k tokens in length. %\sj{Let's briefly introduce the length interpolation/extrapolation and long-llama methods here.}
Benchmarking long-context capability of LLMs involves evaluating on several tasks which naturally have a long context, such as abstractive summarization or multi-document QA. The following benchmarks have been proposed for long-context evaluation of LLMs:

\begin{itemize}[leftmargin=*,topsep=3pt,itemsep=3pt,parsep=0pt]
    \item \textbf{SCROLLS} \citep{scrolls} is a popular evaluation benchmark made of 7 datasets with naturally long input. The tasks cover summarization (GovReport \citep{govreport}, SummScreen \citep{summscreen}, QMSum \citep{QMSum}), question-answering (Qasper \citep{qasper}, NarrativeQA \citep{NarrativeQA}, QuALITY \citep{quality}) and natural language inference (ContractNLI \citep{contractnli}).
    
    \item \textbf{ZeroSCROLLS} \citep{zeroscrolls} builds on SCROLLS (discarding ContractNLI, reusing the 6 other datasets, and adding SQuALITY \citep{SQuALITY}, MuSiQue \citep{musique}, SpaceDigest and BookSumSort) and only considers the zero-shot setting, evaluating LLMs out-of-the-shelf. 

    \item \textbf{LongBench} \citep{longbench} sets a bilingual English/Chinese long-context benchmark of 21 datasets across 6 tasks: single-document QA, multi-document QA, summarization, few-shot learning, synthetic tasks (passage retrieval), and code completion.

    \item \textbf{L-Eval} \citep{leval} re-uses 16 existing datasets and creates 4 datasets from scratch to make a diverse, long-context benchmark, with average length per task above 4k tokens. The authors advocate for LLM judges evaluation (especially GPT-4) rather than n-gram for long-context evaluation.

    \item \textbf{BAMBOO} \citep{bamboo} creates a long-context LLM evaluation benchmark focused on removing pre-training data contamination by collecting only recent data in the evaluation datasets.

    \item \textbf{M4LE} \citep{m4le} introduces a broad-scope benchmark, splitting 36 datasets in 5 understanding abilities: explicit single-span, semantic single-span, explicit multiple-span, semantic multiple-span, and global understanding. 

\end{itemize}

\begin{wraptable}{r}{0.65\textwidth}
    \centering
    \vspace{-1.0em}
    \resizebox{0.65\textwidth}{!}{
        \begin{tabular}{@{}lcccccccccc@{}}
            \toprule
            
            \multirow{2}{*}{\textbf{Model}} 
            & \multicolumn{4}{c}{\textbf{Summarization}}     
            & \multicolumn{4}{c}{\textbf{QA}}
            & \multicolumn{2}{c}{\textbf{Agg}} \\ 
            \cmidrule(l){2-5} 
            \cmidrule(l){6-9} 
            \cmidrule(l){10-11} 
            & \textbf{GR} & \textbf{SS} & \textbf{QM} & \textbf{SQAL} & \textbf{QPER} & \textbf{NAQA} & \textbf{QAL} & \textbf{MQE} & \textbf{SD} & \textbf{BSS} \\ 
            
            \midrule
        
            \textbf{Llama-2-70B-chat + retrieval}   & - & - & 18.3 & - & 31.3 & 24.5 & 69.6 & 26.7 & - & - \\
            \textbf{Llama-2-long-70B-chat}          & \underline{26.0} & 15.0 & \textbf{20.0} & 20.9 & 52.0 & \textbf{31.7} & \underline{82.6} & \underline{27.3} & \underline{55.5} & 46.2 \\ 
             
            \midrule
            
            \textbf{\chat}                          & 21.3 & 16.1 & 15.6 & 20.4 & 49.3 & 25.1 & 66.6 & 27.1 & 49.1 & 49.8 \\
            \textbf{\chat-16k}                      & 24.3 & \underline{16.2} & 17.4 & \underline{21.4} & \underline{50.0} & \underline{29.5} & 72.0 & 27.0 & 54.1 & \underline{54.6} \\
            \textbf{GPT-4}                          & \textbf{26.3} & \textbf{17.3} & \underline{18.5} & \textbf{22.6} & \textbf{50.7} & 27.6 & \textbf{89.2} & \textbf{41.1} & \textbf{62.8} & \textbf{60.5} \\ 
            
            \bottomrule
        \end{tabular}
    }
    \caption{Model performance on ZeroSCROLLS. \textbf{GR} is short for GovReport, \textbf{SS} is SummScreen, \textbf{QM} is QMSum, \textbf{SQAL} is SQuALITY, \textbf{QPER} is Qasper, \textbf{NAQA} is NarrativeQA, \textbf{QAL} is QuALITY, \textbf{MQE} is MuSiQue, \textbf{SD} is SpaceDigest, and \textbf{BSS} is BookSumSort.}
    \vspace{-0.5em}
    \label{tab:longcontext}
\end{wraptable}

\paragraph{Performance of LLMs} On LongBench, L-Eval, BAMBOO and M4LE benchmarks, \chat~or its 16k version largely outperform all open-source LLMs, such as Llama-2 \citep{llama2}, LongChat \citep{longchat}, or Vicuna \citep{vicuna2023} ; showing that it is not trivial to drive up the performance of open-source LLM on long-input tasks.

Llama-2-long \citep{llama2long} continues pre-training of Llama-2 with 400B tokens using a 16k context window (up from 4k in Llama-2). The resulting Llama-2-long-chat-70B outperforms \chat-16k by 37.7 to 36.7 on ZeroSCROLLS. 
Approaches to tackle long-context tasks include context window extension with positional interpolation \citep{chen2023extending}, which involves another (short) round of fine-tuning with longer context window; and retrieval augmentation \citep{rag}, which necessitates access to a retriever to find relevant information. \cite{xu2023retrieval} combine both these seemingly opposite techniques, pushing a Llama-2-70B above \chat-16k on average over 7 long-context tasks (including 4 datasets from ZeroSCROLLS). We refer to \Cref{tab:longcontext} for a synthesis of results on ZeroSCROLLS. 

%% file: Sections/3_5_application_specific_capabilities.tex
\subsection{Application-specific Capabilities}
In this section, we discuss the desired capabilities in LLMs to tackle specific applications.

\subsubsection{Query-focused Summarization}

\paragraph{Benchmarks} Query-focused or aspect-based summarization requires to generate summaries with regard to a fine-grained question or an aspect category. Query-focused datasets include AQuaMuse \citep{AQuaMuSe}, QMSum \citep{QMSum} and SQuALITY \citep{SQuALITY}, while Aspect-based datasets include CovidET \citep{CovidET}, NEWTS \citep{NEWTS}, WikiAsp \citep{WikiAsp}, etc.

\paragraph{Model Performance} \cite{query_focused_summ} finds that standard fine-tuning on training data is still better in performance compared to ChatGPT, with an average of 2 points ROUGE-1 improvement over CovidET, NEWTS, QMSum and SQuALITY.

\subsubsection{Open-ended QA}
\paragraph{Benchmarks} There are two sub-categories in Open-ended QA: either the answers are of short-form or long-form. Short-form datasets include SQuAD 1.1 \citep{SQuAD_1.1}, NewsQA \citep{NewsQA}, TriviaQA \citep{TrivialQA}, SQuAD 2.0 \citep{SQuAD_2}, NarrativeQA \citep{NarrativeQA}, 
Natural Question (NQ) \citep{NaturalQuestions}, Quoref \citep{Quoref} and DROP \citep{drop}. Long-form datasets include ELI5 \citep{ELI5} and doc2dial \citep{doc2dial}. For both short-form and long-form datasets, the evaluation metrics are exact match (EM) and F1 over words in the answers. Answering Open-ended QA requires the model to comprehend the provided context, or retrieve related knowledge if there's no context provided. 

\paragraph{Model Performance} InstructRetro \citep{InstructRetro} shows large improvement over GPT-3 on NQ, TriviaQA, SQuAD 2.0 and DROP, while having 7-10 percent improvement compared to a proprietary GPT-instruct model of similar size, over a range of short-form and long-form open-ended QA datasets. Initialized from a pre-trained GPT model, InstructRetro continues pre-training with retrieval and then undergoes instruction tuning.\footnote{InstructRetro is not yet open-sourced.}

\subsubsection{Medical}
\paragraph{Benchmarks} One desirable capability of LLMs is on contributing medical related tasks to make affordable, high-quality healthcare more accessible to the broader public. 

For mental health, IMHI \citep{MentalLLaMA} benchmark is constructed using 10 existing mental health analysis datasets, including mental health detection: DR \citep{DR}, CLP \citep{CLP}, Dreaddit \citep{Dreaddit}, loneliness, SWMH and T-SID \hphantom{x}  \citep{SWMH}; mental health cause detection: SAD \citep{SAD}, CAMS \citep{CAMS}; mental risk factors detection: MultiWD \citep{MULTIWD}, IRF \citep{IRF}.

For radiology, OpenI \citep{OpenI} dataset and MIMIC-CXR \citep{MIMIC-CXR} datasets both contain radiology reports with findings and impressions text.

\paragraph{Performance of LLMs} For mental health, MentalLlama-chat-13B \citep{MentalLLaMA} finetunes a Llama-chat-13B model on IMHI training set. MentalLlama-chat-13B model with zero-shot prompting outperforms ChatGPT with few-shot prompting or with zero-shot prompting for 9 out of 10 tasks in IMHI. 
\citet{Radiology-Llama2} proposes to fine-tune a Llama checkpoint to generate impression text given radiology report findings. The resulting Radiology-Llama-2 model outperforms ChatGPT and GPT-4 by a large margin on both MIMIC-CXR and OpenI datasets.

\subsubsection{Generating Structured Responses}

Generating formatted responses in accordance with instructions is a core ability to support agentic capabilities or simply reduce the manual efforts in parsing or translating model responses.

\paragraph{Benchmarks} 
Rotowire \citep{Rotowire}, Struc-Bench-Latex and Struc-Bench-HTML \citep{Struc-Bench} are three datasets aiming to generate formatted tables based on text context. Rotowire contains NBA game summaries with corresponding score tables in plain text. Struc-Bench-Latex contains NBA game summaries and tables in Latex format, while Struc-Bench-HTML includes NBA game summaries with tables in HTML format.

\paragraph{Performance of LLMs} 
Struc-Bench \citep{Struc-Bench} fine-tunes a Llama-7B model on structured generation data. The fine-tuned 7B model outperforms ChatGPT on all benchmarks mentioned above.

\subsubsection{Generating Critiques}

\paragraph{Benchmarks} 
One interesting ability of LLMs is providing feedback or critiques to a response for a question. To benchmark such ability, one can use human annotators or GPT-4 as an evaluator to directly rate the critiques. The original questions can come from any dataset of other capabilities mentioned above.

\paragraph{Models} 
Shepherd \citep{Shepherd} is a 7B model initialized from Llama-7B and trained on community collected critique data and 1,317 examples of high quality human annotated data. Shepherd generates critiques on a range of diverse NLP datasets: AlpacaFarm, FairEval, CosmosQA \citep{CosmosQA}, OBQA \citep{OBQA}, PIQA \citep{PIQA}, TruthfulQA and CritiqueEval. With GPT-4 as an evaluator, Shepherd wins or equals ChatGPT over 60\% of the time. With human evaluators, Shepherd is almost on-par with ChatGPT.  

%% file: Sections/3_6_trustworthy.tex
To ensure LLMs can be trusted by humans in real-world applications, an important consideration is their reliability. For example, concerns on hallucination \citep{ye2022unreliability, zhao2023can} and safety~\citep{zhiheng2023safety} could deteriorate user trust in LLMs and lead to risks in high-impact applications.

\subsubsection{Hallucination}

\paragraph{Benchmarks} Various benchmarks have been proposed for better evaluating hallucinations in LLMs. Specifically, they consist of both large-scale datasets, automated metrics, and evaluation models.
\begin{itemize}[leftmargin=*,topsep=3pt,itemsep=3pt,parsep=0pt]
    \item \textbf{TruthfulQA}~\citep{truthfulqa} is a benchmark question-answering (QA) dataset consisting of questions spanning 38 categories. The questions are crafted such that some humans would falsely answer them due to misconceptions.
    \item \textbf{FactualityPrompts}~\citep{lee2022factuality} is a dataset that measures hallucinations for open-ended generation. It consists of factual and non-factual prompts to study the impact of prompts on LLM's continuations. 
    \item \textbf{HaluEval}~\citep{li2023halueval} is a large dataset of generated and human-annotated hallucinated samples. It spans three tasks: question answering, knowledge-grounded dialogue, and text summarization.
    \item \textbf{FACTOR}~\citep{muhlgay2023generating} proposes a scalable approach for evaluating LM factuality: it automatically transforms a factual corpus into a faithfulness evaluation benchmark. The framework is used to create two benchmarks: Wiki-FACTOR and News-FACTOR.
    \item \textbf{KoLA}~\citep{yu2023kola} constructs a Knowledge-oriented LLM Assessment benchmark (KoLA) with three crucial factors: mimicking human cognition for ability modeling, using Wikipedia for data collection, and designing contrastive metrics for automatic hallucination evaluation. 
    \item \textbf{FActScore}~\citep{min2023factscore} proposes a new evaluation that first breaks an LLM's generation into a series of atomic facts, and then computes the percentage of atomic facts supported by a reliable knowledge source.
    \item \textbf{Vectara's Hallucination Evaluation Model}~\citep{vectura} is a small language model that is fine-tuned as a binary classifier to classify a summary as factually consistent (or not) with the source document. Then, it is used to evaluate and benchmark hallucinations of summaries generated by various LLMs.
    \item \textbf{FacTool}~\citep{factool} is a task and domain agnostic framework for detecting factual errors of texts generated by LLMs. 
    \item \textbf{SummEdits}~\citep{laban2023llms} is a 10-domain benchmark that measures LLMs' abilities to detect factual inconsistencies.
    % \sj{Let's include SumEdits?} %added
\end{itemize}

Besides the newly introduced hallucination benchmarks, prior QA datasets based on real-world knowledge are also widely used for measuring faithfulness, such as HotpotQA~\citep{yang2018hotpotqa}, OpenBookQA~\citep{openbookqa}, MedMC-QA~\citep{pal2022medmcqa}, and TriviaQA~\citep{TrivialQA}. 
Besides datasets and automated metrics, human evaluation is also widely adopted as a reliable measure for faithfulness.

\paragraph{Performance of LLMs}

% \begin{wraptable}{r}{0.8\textwidth}
\begin{table}
    \centering
    \resizebox{0.95\textwidth}{!}{
        \begin{tabular}{lcccccc}
            \toprule
            
            \textbf{Models}           
            & \textbf{TruthfulQA} 
            & \textbf{FactScore} 
            & \textbf{HotpotQA} 
            & \textbf{OpenBookQA} 
            & \textbf{MedMC-QA} 
            & \textbf{TriviaQA} \\
            
            \midrule
            
            \textbf{Playtus}                                 & \underline{62.3} & - & - & -  & - & - \\
            \textbf{CoVe + Llama-65B}                        & - & \textbf{71.4} & - & - & - & - \\
            \textbf{CoK + GPT-3.5-turbo}                     & - & - & \underline{35.4} & - & \textbf{73.3} & - \\
            \textbf{CRITIC + GPT-3.5-turbo}                  & - & - & \textbf{38.7} & - & - & 75.1  \\
            \textbf{KSL + GPT-3.5-turbo}                     & - & - & - & \textbf{81.6} & - & - \\
            \textbf{PKG + text-davinci-002}                  & - & - & - & - & \underline{47.4} & - \\
            \textbf{\citet{cohen2023lm} + text-davinci-002}  & - & - & - & - & - & \textbf{83.1} \\
            
            \midrule
            
            \textbf{GPT-3.5-turbo}                           & \textbf{47.0} & \underline{58.7} & 24.0 & \underline{78.3} & 44.4 & \underline{79.3} \\ 
            
            \bottomrule                        
        \end{tabular}
    }
    \caption{Model performance on hallucination benchmarks.}
    \label{tab:halu}
\end{table}
% \end{wraptable}

There already exist a few surveys \citep{zhang2023siren,rawte2023survey} that investigate different proposed methodologies to reduce on hallucination in detail.
Specifically, the methods that surpass the current \chat{} performance can be either incorporated during fine-tuning or only at inference time. Selected model performances on various benchmarks 
% metrics\sj{? what do you mean by metrics?} 
are shown in Table \ref{tab:halu}.

% $\bullet$\textbf{Improving Data Quality During Fine-tuning}\enspace
During fine-tuning, improving data quality in correctness and relevance can lead to less-hallucinated models. 
\citet{lee2023platypus} curated a content-filtered, instruction-tuned dataset, focusing on high-quality data in the STEM domain. A family of LLMs is fine-tuned on this filtered dataset and merged. The resulting family, named Platypus, demonstrates a substantial improvement on TruthfulQA (approximately 20\%) compared to \chat. 

During inference, existing techniques apply specific decoding strategies, external knowledge augmentation, and multi-agent dialogue. 
% $\bullet$\textbf{Improving Decoding Strategies During Inference}\enspace
For decoding, \citet{dhuliawala2023chain} introduce Chain-of-Verification (CoVe), where the LLM drafts verification questions and self-verify the responses, which leads to a substantial improvement on FactScore over \chat. 

% $\bullet$\textbf{Knowledge Augmentation During Inference}\enspace
For external knowledge augmentation, various frameworks incorporate different searching and prompting techniques to improve \chat{} performance
% \sj{revise this...}. %revised 
\citet{li2023chain} design Chain-of-Knowledge (CoK), which retrieves evidences from heterogeneous knowledge sources and synthesizes it before answering. 
% \citet{yu2023improving} proposes REFEED, a pipeline designed to enhance LLMs by providing automatic retrieval feedback in a plug-and-play framework. 
\citet{peng2023check} propose LLM-AUGMENTER, which augments LLMs with a set of plug-and-play modules and iteratively revises LLM prompts to improve model responses using feedback generated by utility functions. Knowledge Solver (KSL)~\citep{feng2023knowledge} tries to teach LLMs to search for essential knowledge from external knowledge bases by harnessing their own strong generalizability. CRITIC~\citep{critic} allows LLM to validate and progressively amend their own outputs in a manner similar to human interaction with tools. \citet{luo2023augmented} introduce a Parametric Knowledge Guiding (PKG) framework, which equips LLMs with a knowledge-guiding module to access relevant knowledge without altering the LLMs’ parameters. These inference techniques then improve answer accuracy compared to the naive prompting strategy using \chat. Currently, \chat{} has also incorporated a retrieval plugin~\citep{chatgptplugins} that accesses external knowledge to reduce hallucinations. 

For multi-agent dialogue, 
\citet{cohen2023lm} facilitate a multi-turn interaction between the Examinee LLM that generated the claim and another Examiner LLM which introduces questions to discover inconsistencies. Through the cross-examination process, performance on various QA tasks is improved. \citet{du2023improving} ask multiple language model instances to propose and debate their individual responses and reasoning processes over multiple rounds to arrive at a common final answer, which improves on multiple benchmarks. 

\subsubsection{Safety}

\paragraph{Benchmarks}
Safety concerns in LLMs can mostly be grouped into three aspects~\citep{zhiheng-etal-2023-safety}: social bias, model robustness, and poisoning issues. To gather datasets that better evaluate the above aspects, several benchmarks have been proposed: 
\begin{itemize}[leftmargin=*,topsep=3pt,itemsep=3pt,parsep=0pt]
    \item \textbf{SafetyBench}~\citep{safetybench} is a dataset which consists of 11,435 diverse multiple choice questions spanning 7 distinct categories of safety concerns.
    \item \textbf{Latent Jailbreak}~\citep{qiu2023latent} introduces a benchmark that assesses both the safety and robustness of LLMs, emphasizing the need for a balanced approach.
    \item \textbf{XSTEST}~\citep{xstest} is a test suite that systematically identifies exaggerated safety behaviors, such as refusing safe prompts. 
    \item \textbf{RED-EVAL}~\citep{redteaming} is a benchmark to perform red-teaming~\citep{ganguli2022red} to conduct safety evaluations of LLMs using a Chain of Utterances (CoU)-based prompt.
\end{itemize}
Besides automated benchmarks, an important measure for safety is human evaluation~\citep{dai2023safe}, where crowdworkers label the responses as safe or harmful. Some studies also attempt to collect such labels from GPT-4, as research shows that it can replace human evaluators in evaluating alignment abilities~\citep{chiang2023can}.

\paragraph{Performance of LLMs}
Based on current evaluation~\citep{safetybench, xstest}, \chat{} and GPT-4 remain at the top for safety evaluations. 
This can be largely attributed to Reinforcement Learning with Human Feedback (RLHF)~\citep{bai2022training}, which first collects a human preference dataset on responses, then trains a reward model to mimic human preferences, and finally uses RL to train the LLM to align with human preferences. In the process, LLMs learn to demonstrate desired behaviors and exclude harmful responses such as impolite or biased answers. 
However, the RLHF procedure requires collecting a large number of expensive human annotations, which hinders its use for open-source LLMs. 
To democratize endeavors on advancing the safety alignment of LLMs, \citet{beavertails} gather a human-preference dataset to disentangle harmlessness and helpfulness from the human-preference score, thus providing separate ranking data for the two metrics. Experiments show that disentangling human preferences enhances safety alignment.
\citet{bai2022constitutional} seek to increase safety with RL from AI Feedback (RLAIF), where the preference model is trained using LLM-generated self-critiques and revisions. Direct Preference Optimization (DPO)~\citep{dpo} reduces the need to learn a reward model and learn from preferences directly with a simple cross-entropy loss. Combining and improving these methodologies could lead to potential improvements in safety for open-source LLMs.

%% file: Sections/4_1_trend.tex
\begin{figure}[t]
  \centering
    \includegraphics[width=1.0\textwidth]{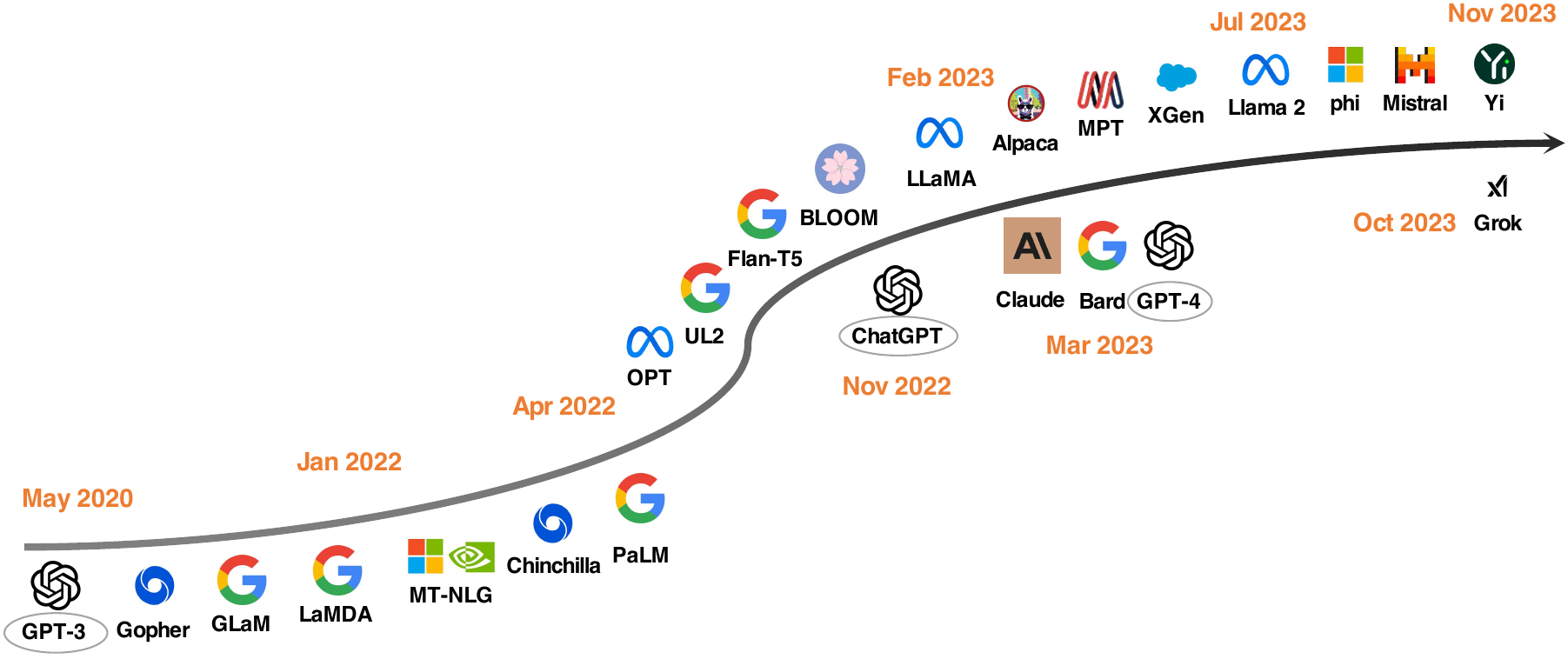}
  \caption{LLM development timeline. The models below the arrow are closed-source while those above the arrow are open-source.}
\end{figure}

Ever since \citet{gpt3} demonstrated that a frozen GPT-3 model can achieve impressive zero- and few-shot performance on a variety of tasks, numerous efforts have been made to advance the development of LLMs. One line of research focused on scaling up model parameters, including Gopher \citep{rae2021scaling}, GLaM \citep{glam}, LaMDA \citep{lamda},  MT-NLG \citep{smith2022using} and PaLM \citep{palm}, culminating at 540B parameters. Despite exhibiting remarkable capabilities, the closed-source nature of these models limited their widespread application, thereby leading to a growing interest in developing open-source LLMs \citep{opt,bloom,survey-llm-zhao}.

Rather than scaling up model size, another line of research explored better strategies or objectives for pre-training smaller models, such as Chinchilla \citep{chinchilla} and UL2 \citep{ul2}. Beyond pre-training, considerable attention has been devoted to studying instruction tuning of LMs, \eg\ FLAN \citep{instructiontuning}, T0 \citep{sanh2021multitask} and Flan-T5 \citep{chung2022scaling}. 
%Flan-PaLM 540B achieves state-of-the-art performance on several benchmarks, such as 75.2% on five-shot MMLU

The emergence of OpenAI's ChatGPT a year ago greatly changed the research focus of the NLP community \citep{qin2023chatgpt}. To catch up with OpenAI, Google and Anthropic introduced Bard and Claude, respectively. While they show comparable performance to ChatGPT on many tasks, there is still a performance gap between them and the latest OpenAI model GPT-4 \citep{gpt4}. As the success of these models is primarily attributed to reinforcement learning from human feedback (RLHF) \citep{ppo,ouyang2022training}, researchers have explored various ways to improve RLHF \citep{yuan2023rrhf,rafailov2023direct,lee2023rlaif}.

%%llama
To promote research of open-source LLMs, Meta released Llama series models \citep{llama,llama2}. Since then, open-source models based on Llama have started to emerge explosively, and some of them have achieved better performance than Llama-2-70B with much fewer parameters, e.g., Yi-34B~\footnote{https://github.com/01-ai/Yi}, Skywork~\citep{wei2023skywork}, Aquila-2~\footnote{https://github.com/FlagAI-Open/Aquila2}, QWen~\citep{qwen}, InternLM~\citep{2023internlm}, Yuan-2.0~\footnote{https://github.com/IEIT-Yuan/Yuan-2.0}, XVERSE~\footnote{https://github.com/xverse-ai/XVERSE-65B}, Mistral~\citep{mistral}, and Mixtral-MoE~\citep{mixtral}. Although some of them do not release the technical details for training, which should not be viewed as fully open-sourced, the publicly available weights can still serve as the stable foundation for future research.
Besides, one another representative research direction is to fine-tune Llama with instruction data. For example, Alpaca \citep{alpaca} is fine-tuned from Llama on 52K instruction-following demonstrations generated by self-instruct \citep{selfinstruct} from OpenAI's text-davinci-003. Vicuna \citep{vicuna2023} is an open-source chatbot fine-tuned from Llama on 70K user-shared conversations collected from ShareGPT. Lima \citep{lima} demonstrates that a small number of carefully annotated instruction samples can be enough to reveal the potential capabilities of LLMs fully. By leveraging Evol-Instruct to use LLMs to generate large amounts of instruction data with varying complexity, WizardLM \citep{wizard-lm} achieves comparable performance to OpenAI's ChatGPT.

The ongoing research has also explored improving the agent \citep{lemur,agenttuning,gorilla,toolllm}, logical reasoning \citep{roziere2023code,wizard-math,wizard-code} and long-context modeling \citep{tworkowski2023focused,llama2long,xu2023retrieval} capabilities of Llama-based open-source LLMs. Besides, rather than developing LLMs based on Llama, many efforts have been devoted to training powerful LLMs from scratch, including generic LLMs (\eg\ MPT \citep{MPT}, Falcon \citep{falcon40b}, XGen \citep{xgen}, Phi \citep{phi-1,phi-15}, Baichuan \citep{yang2023baichuan}, Mistral \citep{mistral}, 
Grok \citep{Grok} and Yi \citep{Yi}) and domain-specific LLMs (\eg\ Bloomberggpt \citep{wu2023bloomberggpt}, FinGPT \citep{yang2023fingpt} and InstructRetro \citep{wang2023instructretro}). We believe that developing more powerful and efficient open-source LLMs to democratize the capabilities of closed-source LLMs should be a quite promising future direction.

%% file: Sections/4_2_summary.tex
% \subsubsection{best open-LLM vs ChatGPT in all domains}
For general capabilities, Llama-2-chat-70B~\citep{llama2} shows improvement over \chat{} in some benchmarks, but remains behind for most others. Zephir-7B~\citep{zephyr} approaches 70B LLMs as a result of distilled direct preference optimization. WizardLM-70B~\citep{wizard-lm} and GodziLLa-70B~\citep{godzila} can achieve comparable performance to \chat{}, which show a promising path forward.

% \subsubsection{In which domains open-LLMs beat/still-fail-to ChatGPT}
% beat
There are also several domains where open-source LLMs are able to beat \chat{}. 
For LLM-based agents, open-source LLMs are able to surpass \chat{} with more extensive and task-specific pre-training and fine-tuning. For example, Lemur-70B-chat~\citep{lemur} performs better in exploring the environment and following feedback on coding tasks. AgentTuning~\citep{agenttuning} improves on unseen agent tasks. ToolLLama~\citep{toolllm} can better grasp tool usage. Gorilla~\citep{gorilla} outperforms GPT-4 on writing API calls.
For logical reasoning, WizardCoder~\citep{wizard-code} and WizardMath~\citep{wizard-math} improve reasoning abilities with enhanced instruction tuning. Lemur~\citep{lemur} and Phi~\citep{phi-1,phi-15} achieve stronger abilities by pre-training on data with higher quality.
For modelling long contexts, Llama-2-long~\citep{llama2long} can improve on selected benchmarks by pre-training with longer tokens and a larger context window. \citet{xu2023retrieval} improves over 7 long-context tasks by combining context window extension with positional interpolation and retrieval augmentation.
For application-specific capabilities, InstructRetro~\citep{InstructRetro} improves on open-ended QA by pre-training with retrieval and instruction tuning. With task-specific fine-tuning, MentaLlama-chat-13B~\citep{MentalLLaMA} outperforms \chat{} in mental health analysis datasets. Radiology-Llama2~\citep{Radiology-Llama2} can improve performance on radiology reports. Stru-Bench~\citep{Struc-Bench}, a fine-tuned 7B model, can improve structured response generation compared to \chat{}, which is a core ability to support agentic tasks. Shepherd~\citep{Shepherd}, with only 7B parameters, can achieve comparable or better performance compared to \chat{} in generating model feedbacks and critiques.
For trustworthy AI, hallucinations can be reduced by fine-tuning with data of higher quality~\citep{lee2023platypus}, context-aware decoding techniques~\citep{dhuliawala2023chain}, external knowledge augmentation such as \citet{li2023chain, yu2023improving, peng2023check, feng2023knowledge}, or multi-agent dialogue~\citep{cohen2023lm, du2023improving}.

There are also domains where \chat{} and GPT-4 remain unbeatable, such as AI safety. Due to the large-scale RLHF~\citep{bai2022training} involved in GPT models, they are known to demonstrate safer and more ethical behaviors, which is probably a more important consideration for commercial LLMs compared to open-source ones. However, with the recent efforts on democratizing the RLHF process~\citep{bai2022constitutional, dpo}, we could expect to see more performance improvements for open-source LLMs in safety.

%% file: Sections/4_3_recipe.tex
Training an LLM involves complex and resource-intensive practices, including data collection and preprocessing, model design, and training process.
While there is a growing trend of releasing open-source LLMs regularly, the detailed practices of the leading models are often kept secret unfortunately.
Below we list some best practices widely acknowledged by the community.

\paragraph{Data}
Pre-training involves the use of trillions of data tokens, often sourced from publicly accessible sources. Ethically, it is crucial to exclude any data that includes personal information of private individuals \citep{llama2}.
Unlike pre-training data, fine-tuning data is smaller in quantity but superior in quality. Fine-tuned LLMs with top-quality data have shown improved performance, particularly in specialized areas \citep{godzila, agenttuning, lemur, wizard-lm}. 

\paragraph{Model Architecture}
While the majority of LLMs utilize the decoder-only transformer architecture, different techniques in the model are employed to optimize efficiency. Llama-2 implements Ghost attention for improved multi-turn dialogue control \citep{llama2}. Mistral \citep{mistral} employs sliding window attention to handle extended context lengths.

\paragraph{Training}
The process of supervised fine-tuning (SFT) with instruction tuning data is vital. For high quality outcomes, tens of thousands of SFT annotations are sufficient, as evidenced by the 27,540 annotations used for Llama-2 \citep{llama2}.
The diversity and quality of these data are essential \citep{wizard-lm}.
In the RLHF stage, proximal policy optimization (PPO) \citep{ppo} is often the preferred algorithm to better align the model's behavior with human preferences and instruction adherence, playing a key role in enhancing LLM safety. An alternative to PPO is direct preference optimization (DPO) \citep{dpo}. Zephyr-7B \citep{zephyr}, for instance, employs distilled DPO and has shown results comparable to 70B-LLMs on various general benchmarks, even surpassing \chat~on AlpacaEval.

\paragraph{Serving (Inference)}
% \sj{include efficient serving.}
Inference in LLMs can be slow. New methods have been created to expedite this process. One such technique, speculative decoding \citep{speculativedecodinggoogle,speculativedecodingdeepmind}, accelerates sampling from autoregressive models. By leveraging approximate model generation and processing multiple tokens simultaneously, this approach can speed up the inference by 2 to 3 times without altering the results.
Additionally, vLLM \citep{vllm} is a high-speed, user-friendly library specifically designed for LLM fast inference. It stands out for its state-of-the-art serving capacity and effective memory utilization.

%% file: Sections/4_4_problems.tex
\paragraph{Data Contamination during Pre-training}

% Data contamination is becoming more severe with more released foundation models hiding the source of their pre-training corpus, which can further lead to biased impression over the true generalization of LLMs. Despite the extreme case by manually mixing the the benchmark data into the training set with annotation from human experts or larger models, the data contamination problem is mainly due to the collecting source of benchmark data has already been covered in the pre-training corpus. Although the models are not pre-trained via supervised data on purpose, they can still recover the exact knowledge. As a result, detecting the pre-training corpus of LLMs~\citep{pretrain-data-detect}, investigating the overlap between existing benchmark and widely used pre-training corpus, and measuring the over-fitting to benchmark~\citep{skywork} are all important directions to make LLMs more faithful and reliable. 
The issue of data contamination has become increasingly pronounced with the release of foundation models that obscure the source of their pre-training corpus. This lack of transparency can result in biased perceptions regarding the genuine generalization capabilities of Large Language Models (LLMs). Ignoring the cases where benchmark data is manually integrated into the training set with annotations from human experts or larger models, the root of the data contamination problem lies in the fact that the collecting source of benchmark data is already encompassed in the pre-training corpus. While these models are not intentionally pre-trained using supervised data, they can still acquire exact knowledge. Consequently, it is crucial to address the challenge of detecting the pre-training corpus of LLMs~\citep{pretrain-data-detect}, exploring the overlap between existing benchmarks and widely-used pre-training corpus, and assessing overfitting to benchmarks~\citep{skywork}. These efforts are essential for enhancing the faithfulness and reliability of LLMs. Looking ahead, future directions could involve establishing standardized practices for disclosing pre-training corpus details and developing methods to mitigate data contamination throughout the model development lifecycle.

\paragraph{Close-sourced Development of Alignment}
The application of Reinforcement Learning from Human Feedback (RLHF) for alignment using general preference data has obtained increasing attention within the community. However, only a limited number of open-source LLMs have been augmented with RLHF or alike (DPO) for alignment, primarily due to the scarcity of high-quality, publicly available preference datasets and pre-trained reward models. Some initiatives~\citep{bai2022training, qa-feedback, ultra-feedback} have sought to contribute to the open-source community. Yet, we are still facing the challenges lacking diverse, high-quality and scalable preference data in complex reasoning, programming, and safety scenarios.

\paragraph{Difficulty in Continuous Improvements over Fundamental Abilities}

% Review the breakthrough in fundamental abilities summarized in this paper, and we will find the scenarios a bit embarrassing: (1) Tremendous efforts are made to explore better data mixture during pre-training to provide more balanced, 
 % and robust abilities in building stronger foundation models. Yet, the cost of exploration often makes this direction unapproachable. (2) Models surpassing \chat~or GPT-4 are mostly based on knowledge distillation from closed-source models and more expert annotation. Though efficient, heavy reliance on knowledge distillation will also conceal the potential problems that whether the proposed approaches really make effects when being scaled to the teacher model. Besides, LLMs are expected to play the role of agent, and provide reasonable interpretations to support the decisions. As a result, data annotation is also becoming expensive in order to make LLMs applicable to real scenarios. In a nutshell, optimization through either knowledge distillation or expert annotation cannot realize continuous improving, and will soon approaching the upper bound.
Reviewing the breakthroughs in fundamental abilities outlined in this paper reveals somewhat challenging scenarios: (1) Considerable efforts have been invested in exploring improved data mixtures during pre-training to enhance balance and robustness in constructing more potent foundation models. However, the associated exploration costs often render this approach impractical. (2) Models surpassing \chat~or GPT-4 are predominantly based on knowledge distillation from closed-source models and additional expert annotation. While efficient, heavy reliance on knowledge distillation may mask potential issues concerning the effectiveness of proposed approaches when being scaled to the teacher model. Moreover, LLMs are anticipated to act as agents and provide reasonable interpretations to support decisions, while annotating the agent-style data to make LLMs applicable to real-world scenarios is also expensive and time-consuming. In essence, optimization through knowledge distillation or expert annotation alone cannot realize continuous improvement and is likely to approach an upper bound. Future research directions may involve exploring novel methodologies, such as unsupervised or self-supervised learning paradigms, to enable continuous advancements in fundamental LLM abilities while mitigating the associated challenges and costs.